\documentclass[twoside,11pt]{article}

\usepackage{tabularx}
\usepackage{jmlr2e}

\usepackage{url}            
\usepackage{booktabs}       
\usepackage{amsfonts}       

\usepackage{wrapfig}
\usepackage{mystyle}
\usepackage{enumitem}

\usepackage{tikz}
\usetikzlibrary{calc,matrix,decorations.markings,decorations.pathreplacing}

\usepackage{xcolor}

\jmlrheading{x}{2018}{x-xx}{x/xx}{xx/xx}{Jean Kossaifi, Yannis Panagakis, Anima Anandkumar and Maja Pantic}

\title{TensorLy: Tensor Learning in Python}

\editor{}

\author{\name Jean Kossaifi\(^{1}\) \email jean.kossaifi@imperial.ac.uk\\ 
\name Yannis Panagakis\(^{1,2}\) \email i.panagakis@imperial.ac.uk\\ 
\name Anima Anandkumar\(^{3, 4}\) \email anima@amazon.com\\
\name Maja Pantic\(^{1}\) \email m.pantic@imperial.ac.uk\\ 
\addr \(^1\)Imperial College London
\hfill \addr \(^{2}\)Middlesex University,\\
\addr \(^{3}\) Amazon AI
\hfill \addr \(^{4}\) California Institute of Technology
} 

\begin{document}

\maketitle

\begin{abstract}Tensors are higher-order extensions of matrices. While matrix methods form the cornerstone of machine learning and data analysis, tensor methods have been gaining increasing traction. However,  software support for tensor operations is not on the same footing. In order to bridge this gap, we have developed \emph{TensorLy}, a high-level API for tensor methods and deep tensorized neural networks in Python. TensorLy aims to follow the same standards adopted by the main projects of the Python scientific community, and seamlessly integrates with them. Its BSD license makes it suitable for both academic and commercial applications. 
TensorLy's backend system allows users to perform computations with NumPy, MXNet, PyTorch, TensorFlow and CuPy. They can be scaled on multiple CPU or GPU machines. In addition, using the deep-learning frameworks as backend allows users to easily design and train deep tensorized neural networks. TensorLy is available at \url{https://github.com/tensorly/tensorly}

\end{abstract}

\section{Introduction}

\paragraph{}
\emph{Tensors} are higher-order extensions of matrices. While matrices are indexed by two indices (and hence, order-2 tensors), tensors can be indexed by an arbitrary number of indices. Tensors have a rich history, stretching almost a century, and have been used in diverse fields such as psychometrics, quantum systems, signal processing etc.  Only recently, tensors have been employed in machine learning and data analytics. This can be attributed to the availability of large-scale multi-dimensional and multi-modal datasets. Tensors form the natural framework to encode and operate on such data structures.

For multi-dimensional datasets, there are natural extensions of traditional dimensionality-reduction methods, such as principal component analysis (PCA), to higher dimensions. These involve tensor decompositions and have been applied in a number of areas, including the theoretical analysis of deep neural nets (\cite{expressive_deep_tensor}). Another line of work aims to operate on higher-order moments of the data distribution, i.e. beyond pairwise correlations. This is shown to be fruitful for learning a wide range of probabilistic latent-variable models such as Gaussian mixtures, Independent Component Analysis, topic models etc~(\cite{decomposition_latent_anandkumar}). More recently, deep tensorized neural networks have been explored for a variety of applications. They extend linear-algebraic operations in various layers to tensor algebraic operations, such as tensor contraction and tensor regression, in convolutional architectures~(\cite{novikov2015tensorizing,trl}), generative adversarial networks (\cite{tensor-gan}) and sequence models~(\cite{tensor-rnn}).

Interested readers are referred to several surveys on this topic. Some works focus on the basics of multi-linear (tensor) algebra and different types of tensor decompositions, \cite{tensor_decomposition_kolda,tensor_decomposition_sidiropoulos}. 
Others focus on algorithmic advances, \cite{nonnegative_cichocki_2009,lu_2011,survey_grasedyck_2013,cichocki_2015,tensor_mining_papalexakis}.
Recent surveys focus on their applications (\cite{acar_yener_2009}) and uses in learning probabilistic models (\cite{monographtensors}).

Thus, tensor methods can have a profound impact on data analytics and machine learning with clear theoretical, algorithmic, and practical advantages over their matrix counterparts. However, as opposed to matrix methods, tensor methods have not yet been widely adopted by data scientists. This can be mainly attributed to the lack of available software libraries for tensor operations and decompositions, in programming languages that data scientists and practitioners are familiar with (e.g., Python, Java, Scala, etc).

Even though some tensor libraries exist, they are implemented in non-free platforms (e.g., MATLAB's TensorToobox,
\cite{tensor_toolbox_kolda} and TensorLab, \cite{tensorlab_lathauwer}) or in low-level languages like C++ (e.g., TH++). Python is emerging as a language of choice for machine learning, as witnessed with the success of scikit-learn (\cite{scikit-learn}), and is increasingly used in both academic and industrial research projects. However, there is not yet a Python library implementing tensor operations, decomposition, and learning. The existing ones (e.g., scikit-tensor) offer only limited algorithms (e.g., decomposition only) and/or have restrictive licenses. Moreover, widely-adopted deep-learning libraries such as Tensorflow and Torch lack advanced tensor operations. For applications to data analytics, machine learning and deep learning, there is an urgent need for well-developed and documented open-source libraries that include methods for tensor decompositions.

In this paper, to address the aforementioned need, the \emph{TensorLy}~\footnote{\emph{TensorLy}'s source code available at \url{https://github.com/tensorly/tensorly} and documentation at \url{http://tensorly.org}} library is introduced, allowing several contributions over the existing libraries for tensor methods. In particular, TensorLy
a) provides state-of-the-art tensor learning, including core tensor operations and algebra, tensor decomposition and tensor regression methods;
b) has a flexible \emph{backend} system that allows switching between NumPy, MXNet, PyTorch, TensorFlow, and CuPy to perform the computation, and to easily combine tensor methods with deep learning;
c) is open source and BSD licensed;
d) depends by default exclusively on NumPy and SciPy; and
e) is accompanied by extensive tests and documentation.

\section{\emph{TensorLy}: functionalities and implementation}
\label{tensorly}

\paragraph{}
\emph{TensorLy} has been developed with the goal of making tensor learning more accessible and to allow for seamless integration with the python scientific environment. 
To enable algorithms to be easily run at scale, on various hardware, and to combine tensor methods with deep learning, TensorLy has a flexible backend system. This allows tensor operations to be ran using NumPy (\cite{numpy}), MXNet (\cite{mxnet}), PyTorch (\cite{pytorch}), Chainer's CuPy (\cite{chainer}) or TensorFlow with eager execution (\cite{tensorflow}).  \textbf{NumPy} is the standard library for numerical computation in Python. It offers high performance structures for manipulating multi-dimensional arrays.
\textbf{CuPy} works as a drop-in replacement for NumPy with GPU support.
\textbf{MXNet} is a flexible deep learning library with an NDArray structure that allows to efficiently run code on CPU, GPU and multi-machines.
Similarly, \textbf{PyTorch} provides a NumPy-like API with GPU acceleration and auto-grad using dynamical graphs.
\textbf{TensorFlow} is an established machine-learning framework that provides an imperative approach using eager execution. 
Finally, \textbf{ Matplotlib} (\cite{matplotlib}) is a cross-compatible 2D graphics package offering high quality image and graphics generation.

\begin{figure*}
\resizebox{1.01\linewidth}{!}{%
	\input{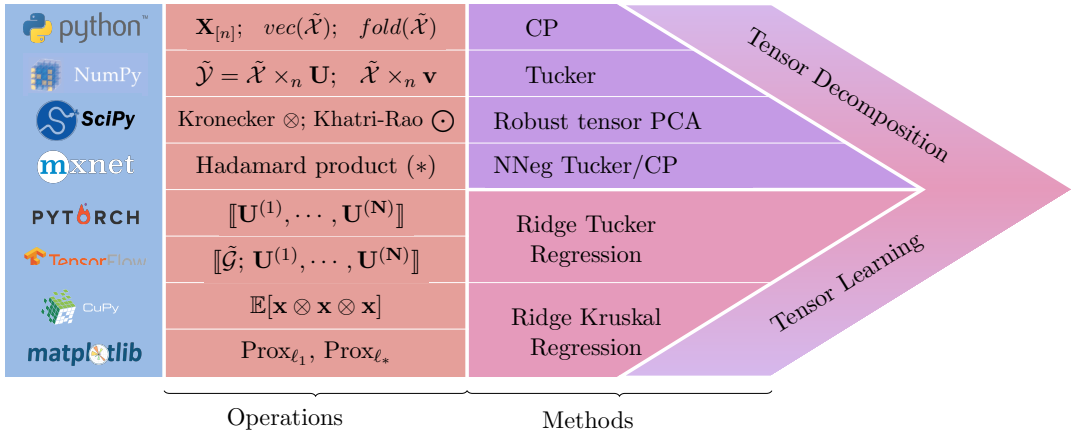}
    \label{tensorly_graph}
}%
	\vspace*{-10pt}
    \caption{TensorLy builds on top of the Python ecosystem and implements Tensor Algebra \emph{Operations}. These tensor operators are then used for higher level \emph{Methods} such as tensor regression or decomposition.}
   \vspace*{-20pt}
\end{figure*}

We aim at making the \emph{API} simple and efficient, following that of scikit-learn, where possible. 
However, while scikit-learn
works with observations (samples) represented as vectors, this library focuses on higher order arrays.
\emph{TensorLy}'s main functionalities in term of tensor operations are summarized in Fig.~\ref{tensorly_graph}, where in the operations column the mathematical notation of \cite{tensor_decomposition_kolda} is adopted. In the methods column, we summarize the implemented core tensor decomposition and regression methods. These include CP and 
Tucker decompositions,
their non-negative versions, 
Robust Tensor PCA, (\cite{robust_goldfarb}) and low-rank Tensor (Kruskal and Tucker) Regression. 
Additionally, using the MXNet, PyTorch, TensorFlow and CuPy backends, it is easy to combine tensor methods and Deep Learning.

We emphasize code quality and ease of utilization for the end user. To that extent, both testing and documentation are an essential part of the package: all functions come with documentation and unit-tests (at the time of writing, the coverage is of 97 \%).

\section{Performance}
\label{experiments}

\paragraph{}
\emph{TensorLy} has been tailored for the Python ecosystem: tensors are multi-dimensional arrays which are effectively manipulated directly by the various methods, decomposition or regression. This allows for competitive performance even though the library is implemented in a high-level, interactive language. The operations are also optimized for speed: tensor operations have been redefined when possible to allow for better performance. In particular, we propose an efficient unfolding of tensors which differs from the traditional one \cite{tensor_decomposition_kolda} by the ordering of the fibers.

Given a tensor,
\( \mytensor{X} \in \myR^{I_1 \times I_2 \times \cdots \times I_N}\), the
mode-\(n\) unfolding of \(\mytensor{X}\) is a matrix \(\mymatrix{X}_{[n]} \in \myR^{I_n,
I_M}\), and is defined by
the mapping from element \( (i_1, i_2, \cdots, i_N)\) to \((i_n, j)\),
\vspace*{-10pt}
$$
\text{with } 
j = \sum_{\substack{k=1,\\k \neq n}}^N i_k 
\times
\prod_{\substack{m=k+1,\\k \neq n}}^N I_m
\text{,\quad and }
M = \prod_{\substack{k=1,\\k \neq n}}^N I_k
\vspace*{-5pt}
$$
This formulation both achieves better performance when using C-ordering of the elements (Numpy and most Python libraries' default), and translates into more natural properties.

\begin{figure}[h!]
  \centering
  \includegraphics[width=0.9\linewidth]{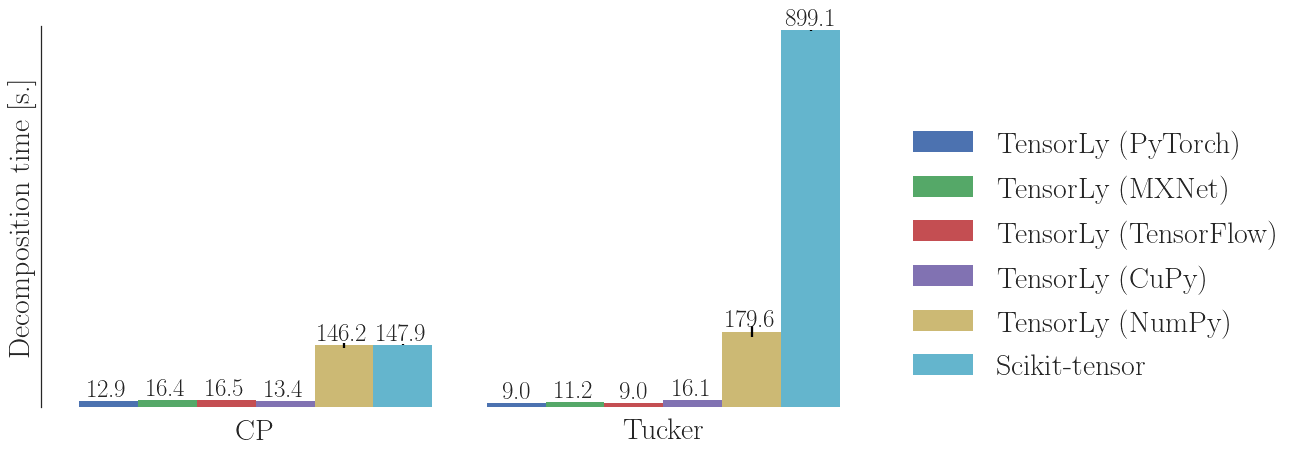}
  \vspace*{-10pt}
  \caption{Speed comparison for Tucker and CANDECOMP-PARAFAC decomposition.}
  \label{comparison}
\end{figure}

\paragraph{Numerical test:} We generated random third order tensors of size \( 500 \times 500 \times 500 \) (125 million elements). We then compared the decomposition speed for a rank--50 CANDECOMP-PARAFAC (CP) and rank $(50, 50, 50)$--Tucker decomposition with TensorLy on CPU (NumPy backend) and TensorLy on GPU (MXNet, PyTorch, TensorFlow and CuPy backends), and Scikit-Tensor (\emph{Sktensor}), Fig.~\ref{comparison}. In all cases we fixed the number of iterations to $100$ to allow for a fair comparison. The experiment was repeated \(10\) times, with the main bar representing the average CPU time and the tip on the bar the standard deviation of the runs. As can be observed, our library offers competitive speed, thanks to optimized formulation and implementation. Experiments were done on an Amazon Web Services p3 instance, with a NVIDIA VOLTA V100 GPU and 8 Intel Xeon E5 (Broadwell) processors.

\vspace*{-10pt}
\section{Conclusion}
\label{conclusion}

\paragraph{}
\emph{TensorLy} makes tensor learning accessible and straightforward by offering state-of-the-art tensor methods and operations through simple and consistent interfaces, under a permissive license. It is optimized to be fast and robust, with systematic unit-tests and documentation. The library's speed and ease of use allow for an efficient comparison of existing methods and easy implementation of new ones. Its flexible backend system allows to transparently switch between libraries and platforms and help combine tensor methods with deep learning. Going forward, we will further extend the available methods with other state-of-the-art methods such as PARAFAC2, DEDICOM and tensor networks (e.g. MPS, also known as tensor-train). We will also investigate extending BLAS primitives for further improvements in performance, as exemplified by \cite{shi2016tensor}.

\small{

}

\end{document}